\documentclass{article} 
\usepackage{iclr2020_conference,times}

\usepackage{amsmath,amsfonts,bm}

\def\eqref#1{equation~\ref{#1}}

\def\1{\bm{1}}

\DeclareMathAlphabet{\mathsfit}{\encodingdefault}{\sfdefault}{m}{sl}
\SetMathAlphabet{\mathsfit}{bold}{\encodingdefault}{\sfdefault}{bx}{n}

\usepackage{hyperref}
\usepackage{url}
\usepackage{subfigure}
\usepackage{amssymb}
\usepackage{amsmath}
\usepackage{booktabs} 
\usepackage{algorithm}
\usepackage{algorithmic}
\usepackage{setspace}
\usepackage{comment}
\usepackage{hyperref}
\usepackage{graphicx}
\usepackage{algorithm}
\usepackage{algorithmic}
\usepackage{setspace}
\usepackage{comment}
\usepackage{microtype}
\usepackage{graphicx}
\usepackage{bbm}
\usepackage{nicefrac}
\usepackage{physics}
\usepackage{mathtools}

\title{Matching Distributions via Optimal Transport for Semi-Supervised Learning}

\author{Fariborz Taherkhani, Hadi Kazemi, Ali Dabouei, Jeremy Dawson, Nasser M. Nasrabadi \\
Lane Department of Computer Science and Electrical Engineering\\
West Virginia University\\
\texttt{\footnotesize $\{$ft0009, hakazemi, ad0046$\}$@mix.wvu.edu, $\{$jeremy.dawson, nasser.nasrabadi$\}$ @mail.wvu.edu}
}
\iclrfinalcopy 
\begin{document}

\maketitle

\begin{abstract}
Semi-Supervised Learning (SSL) approaches have been an influential framework for the usage of unlabeled data when there is not a sufficient amount of labeled data available over the course of training. SSL methods based on Convolutional Neural Networks (CNNs) have recently provided successful results on standard benchmark tasks such as image classification. In this work, we consider the general setting of  SSL problem where the labeled and unlabeled data  come from the same underlying probability distribution. We  propose a new approach that adopts  an Optimal Transport (OT) technique serving as a metric of similarity between discrete empirical probability measures to  provide pseudo-labels for the unlabeled data, which can then be used in conjunction with the initial labeled data to train the CNN model in an SSL manner. We have evaluated and compared our proposed method with state-of-the-art SSL algorithms on standard datasets to demonstrate the superiority and effectiveness of our  SSL algorithm.
\end{abstract}
\section{Introduction}

Recent developments in CNNs have provided promising results for many applications in machine learning and computer vision such as facial recognition \cite{taherkhani2018deep,dabouei2020boosting,taherkhani2020deep,taherkhani2018facial,taherkhani2020pf,kazemi2019identity}, image retrieval \cite{taherkhani2018facial,talreja2018using,taherkhani2020error,kazemi2018unsupervised1}, image generation \cite{kazemi2018unsupervised,kazemi2020preference,kazemi2018unsupervised}, and adversarial attack \cite{dabouei2020smoothfool,dabouei2020exploiting}.  However, the success of CNN models requires  a vast  amount  of well-annotated training data, which is not always feasible to perform manually \cite{NIPS2012_4824,taherkhani2019weakly}. There are essentially two different solutions that are usually used to deal with this problem: 1) Transfer Learning (TL) and 2) Semi-Supervised Learning (SSL). In TL methods \cite{tan2018survey}, the learning of a new task is improved by transferring  knowledge from a related task which has already been learned. SSL methods \cite{oliver2018realistic}, however, tend to learn discriminative models that can make use of the information from an input distribution that is given by a large amount of unlabeled data. To make  use of unlabeled data, it is presumed that the underlying distribution of data has some structure. SSL algorithms make use of at least one of the following structural assumptions: continuity, cluster, or manifold \cite{chapelle2009semi}. In the continuity assumption, data which are close to each other  are more likely to belong to the same class. In the cluster assumption, data tends to form discrete clusters, and data in the same cluster are more likely to share the same label. In modern high-dimensional settings which are typical in signal processing \cite{taherkhani2017restoring,taherkhani2013permutation} or genomics \cite{mohamadi2017arima,mohamadi2019detection,mohamadi2020}, low-dimensional representation of the data are difficult. For example, in parallel with machine learning advances, Mohamadi et al \cite{mohamadi2017arima} present a new trend in signal processing on non-stationary data, where jointly model the linear and non-linear characteristics of time series. In their work as a pioneer work, they discuss how joint linear and non-linear of modeling by ARIMA-GARCH modeling allows accurate prediction of of biomedical time series. They also extend their framework to other interesting application such as genomic \cite{mohamadi2020} which prove the capability of this pioneer framework. However, in the manifold assumption, data lies approximately on a manifold of much lower dimension than the input space which can be classified by using distances and densities defined on the manifold.  Thus, to define a natural similarity distance or divergence between probability measures on a manifold, it is important to consider the geometrical structures of the metric space in which the manifold exists \cite{bronstein2017geometric}.

There are two principal directions that model geometrical structures underlying  the manifold on which the discrete probability measures lie \cite{amari2018information}. The first direction is based on the principal of invariance, which relies on the criterion that the geometry between probability measures should be invariant under invertible transformations of random variables. This perspective is the foundation of the theory of information geometry, which operates as a base for the statistical inference \cite{amari2016information}. The second direction is established by the theory of Optimal Transport (OT), which exploits prior geometric knowledge on the base space in which random variables are valued \cite{villani2008optimal}. Computing OT or Wasserstein distance  between  two random variables equals to achieving a coupling between these two variables that is optimal in the sense that the expectation of the transportation cost between the first and  second variables is minimal. The Wasserstein distance between two probability measures considers the metric properties of the base space on which a structure or a pattern is defined. However, traditional information-theoretic divergences such as the Hellinger divergence and the Kullback-Leibler (KL) divergence are not able to properly capture the geometry of the base space. Thus, the Wasserstein distance is useful for the applications where the structure or geometry of the base space plays a significant role \cite{amari2007methods}. In this work, similar to other SSL methods,  we make a structural assumption about the data in which the data are represented by a CNN model. Inspired by the Wasserstein distance, which exploits properly the geometry of the base space to provide a natural notion of similarity between the discrete empirical measures, we use it to provide pseudo-labels for the unlabeled data to train a CNN model in an SSL fashion. Specifically, in our SSL method, labeled data belonging to each class is a discrete measure.  Thus, all the labeled data create a measure of measures and similarly, the pool of unlabeled data  is also a measure of measures constructed by data belonging to different classes.  Thus, we design a measure of measures OT plan serving as a similarity metric between discrete empirical measures to map the unlabeled measures to the labeled measures based on which, the pseudo-labels for the unlabeled data are inferred. Our SSL method is based on the role of Wasserstein distances in the hierarchical modeling \cite{nguyen2016borrowing}. 
It stems from the fact that the labeled and unlabeled datasets hierarchically create a measure of measures in which each measure is constructed by the data belonging to the same  class.

Computing the exact Wasserstein distance, however, is  computationally expensive and usually is solved by a linear program (Appendix A and D ).  \cite{cuturi2013sinkhorn}  introduced an interesting method which relaxes the OT problem using the entropy of the solution as a strong convex regularizer. The entropic regularization provides two main advantageous: 1) The regularized OT problem relies on Sinkhorn’s algorithm \cite{sinkhorn1964relationship} that is faster by several orders of magnitude than the exact solution of the linear program. 2)  In contrast to exact OT, the regularized OT is a differentiable function
of their inputs, even when the OT problem is used for discrete measures. These advantages have caused that the regularized  OT to receive a lot of attention in machine learning applications such as generating data  \cite{arjovsky2017wasserstein,gulrajani2017improved}, designing loss function \cite{frogner2015learning}, domain adaptation \cite{damodaran2018deepjdot,courty2017optimal}, clustering \cite{cuturi2014fast,ho2017multilevel, mi2018variational} and low-rank approximation \cite{seguy2015principal}.
\section{Related Work}

\textbf{Pseudo-Labeling} is a simple approach whereby a model incorporates it's own predictions on unlabeled data to obtain additional information during the training \cite{rosenberg2005semi, lee2013pseudo, rasmus2015semi,taherkhani2019matrix}. The main downside of these methods is that they are unable to correct their own mistakes where predictions of the model on  unlabeled data are confident but incorrect. In such a case,  the erroneous data not only can not contribute to the training, but the error of the models is amplified during the training as well. This effect is aggravated where the domain of the unlabeled data is different from that of labeled data.
Note that pseudo-labeling in \cite{lee2013pseudo}  is  similar
to entropy regularization \cite{pereyra2017regularizing}, in the sense that it forces the model to provide higher confidence predictions for unlabeled data. However, it differs because it only forces these criteria
on data  which have a low entropy prediction due to the threshold of confidence.

\textbf{Consistency Regularization} can be considered as a way of using unlabeled data to explore a smooth manifold on which all of the data points are embedded \cite{belkin2006manifold}. This simple criterion has provided a set of methods that are currently considered as state of the art for the SSL challenge. Some of these methods are stochastic perturbations \cite{sajjadi2016regularization}, $\pi$-model \cite{laine2016temporal}, mean teacher \cite{tarvainen2017mean}, and Virtual Adversarial Training (VAT) \cite{miyato2018virtual}. The original idea behind stochastic perturbations and  $\pi$-model was first introduced in \cite{bachman2014learning} and has been referred to as pseudo-ensembles. The pseudo-ensembles regularization techniques  are usually designed such that the prediction of the model ideally should not change significantly if the data given to the model is perturbed; in other words, under realistic
perturbations of a data point $x$ ($x \rightarrow x'$), output of the model $f_{\theta}(x)$ should not change significantly.  This goal is achieved by adding 
a weighted loss term such as $d(f_{\theta}(x), f_{\theta}(x'))$ to the total loss of the model $f_{\theta}(x)$, where $d(.,.)$ is mean squared error or Kullback-Leibler divergence which measures a distance between  outputs of the prediction function. The main problem of pseudo-ensemble methods, including  $\pi$-model is that they rely on a potentially unstable target prediction, which can immediately change during the training. 

To address this problem, two methods, including temporal ensembling \cite{laine2016temporal} and mean teacher \cite{tarvainen2017mean}, were proposed to obtain a more stable target output $f'_{\theta}(x)$. Specifically, temporal ensembling uses an exponentially accumulated average of outputs,
$f_{\theta}(x)$, to make the target output smooth and consistent. Inspired by this method, mean teacher instead uses a prediction function which is parametrized by an exponentially accumulated average of $\theta$ during the training.
Like the $\pi$-model, mean teacher adds a mean squared error loss $d(f_{\theta}(x), f'_{\theta}(x))$  as a regularization term to the total loss function for training the network. It has been shown that mean teacher outperforms temporal ensembling in practice \cite{tarvainen2017mean}. Contrary to stochastic perturbation methods which rely on constructing $f_{\theta}(x)$ stochastically, VAT in the first step approximates a small perturbation $r$ to add it to $x$ which significantly changes the
prediction of  the  model $f_{\theta}(x)$. In the next step,  a consistency regularization technique is applied to minimize $d(f_{\theta}(x), f_{\theta}(x+r))$ with respect to   $\theta$ which is the parameters of the model.

\textbf{Entropy Minimization} methods use a loss term which is applied on the  unlabeled data to force  the model $f_{\theta}(x)$ to produce confident predictions (i.e., low-entropy) for all of the samples, regardless of what the actual labels are \cite{grandvalet2005semi}. For example, by assuming the softmax layer of a CNN has $c$ outputs, the loss term applied on  unlabeled data is as follows: $-\sum_{i=1}^c f^{(i)}_{\theta}(x) \log f^{(i)}_{\theta}(x) $. Ideally, this class of methods penalizes the decision boundary that passes near the data points, while they instead force the model to provide a high-confidence prediction \cite{grandvalet2005semi}. It has been shown that entropy minimization on its own, can not produce competitive results \cite{sajjadi2016mutual}. However, entropy minimization can be used in conjunction with  VAT (i.e., EntMin VAT) to provide state of the art
results  in which VAT  assumes a fixed virtual label prediction in the regularization $d({f_{\theta}}(x), f_{\theta}(x+r))$ \cite{miyato2018virtual}. 


\section{Measure of Measures OT}

For any subset $\theta \subset  \mathbb{R}^c$, assume that  ${S}(\theta)$ represents the space of Borel probability measures on $\theta$. The Wasserstein space
of order $k \in [1, \infty)$ of probability measures on $\theta$ is defined as follows: ${S}_k(\theta)=\{ \mathcal{F} \in {S}(\theta): \int ||x||^k d \mathcal{F}(x) < \infty \}$, where, $||.||$ is the Euclidean distance in $\mathbb{R}^c$. Let ${\Pi} (\mathcal{P}, \mathcal{Q})$ denote the set of all probability measures on
$\theta \times \theta$ which have marginals $\mathcal{P}$ and $\mathcal{Q}$; then the $k$-th Wasserstein distance between  $\mathcal{P}$ and $\mathcal{Q}$ in ${S}_k(\theta)$, is defined  as follows \cite{villani2008optimal}: 
\begin{equation}
    W_k(\mathcal{P}, \mathcal{Q})=\bigg(\inf_{\pi \in {\Pi (\mathcal{P}, \mathcal{Q})}} \int_ {\theta^2} ||x-x'||^k d\pi(x,x') \bigg)^{{1}/{k}},
\end{equation}

where  $x\sim \mathcal{P}$,  $x'\sim \mathcal{Q}$ and $k \geq 1$. Explicitly, $W_k(\mathcal{P},
\mathcal{Q})$ is the optimal cost of moving mass from $\mathcal{P}$ to $\mathcal{Q}$, where the cost of moving mass is proportional to the Euclidean distance raised to the power $k$.

In Eq. (1), the Wasserstein  between two probability measures was defined. However,  using a recursion of concepts, we can talk about measure of  measures in which a cloud of measures ($\mathcal{M}'$) is transported to another cloud of measures ($\mathcal{M}$). We define a relevant distance metric on this abstract space as follows: let the space of Borel measures on ${S}_k(\theta)$ be represented by $S_k(S_k(\theta))$; this space is also a Polish, complete and separable metric space as $S_k(\theta)$ is a Polish space (cf. section. 3 in \cite{nguyen2016borrowing}). It will be endowed with a Wasserstein metric $W'_k(.)$ of order $k$ that is induced by a metric $W_k(.)$ on $S_k(\theta)$
as follows:  for any $\mathcal{M}' \in S_k(S_k(\theta))$ and $\mathcal{{M}} \in  S_k(S_k(\theta))$
\begin{equation}
    W'_k(\mathcal{M}', \mathcal{M})=\bigg(\inf_{\pi \in {\Pi (\mathcal{M}',\mathcal{M})}} \int_ {\mathcal{P}_k(\theta) \times \mathcal{P}_k(\theta)} W_k^k(\mathcal{Q},\mathcal{P}) d\pi(\mathcal{Q},\mathcal{P}) \bigg)^{{1}/{k}},
\end{equation}
where, $\mathcal{Q}\sim \mathcal{M}'$, $\mathcal{P} \sim \mathcal{M}$, and $ \Pi (\mathcal{M}', \mathcal{M})$ is the set of all probability
measures on $S_k(\theta) \times S_k(\theta)$ that have marginals $\mathcal{M}'$ and $\mathcal{M}$. Note that the existence of an  optimal solution, $ \pi \in {\Pi (\mathcal{M}',\mathcal{M})}$, is always guaranteed (Appendix E). In words, $W'_k(\mathcal{M}', \mathcal{M})$
corresponds to the optimal cost of transporting mass from $\mathcal{M}'$ to $\mathcal{M}$
, where the cost of moving unit mass in its space of support, $S_k(\theta)$, is proportional to the power
$k$ of the Wasserstein distance $W_k(.)$  in $S_k(\theta)$. 

\section{Matching Measures via  Measure of Measures OT for SSL} 
The goal of our algorithm is to use OT to provide pseudo-labels for the unlabeled data to train a CNN model in an SSL manner. The basic premise in our algorithm is that the discrepancy between two discrete empirical measures which come from the same underlying  distribution is expected to be less than the case where these measures come from two different distributions. In this work, since we make a structural assumption about the data and assume that the labeled and unlabeled data belonging to the same class come from the same distribution (i.e., general setting in SSL), we leverage OT metric to map similar measures from two measure of measures. This is because OT exploits well the structure or geometry of the underlying metric space to provide a natural notion of similarity between empirical measures in the metric space. Here, labeled data belonging to the same class is a measure. Thus, all the initially labeled data construct a measure of measures and similarly, all the unlabeled data is also a  measure of measures constructed by data from different classes. Thus,  we design a  measure of measures OT plan to map the unlabeled measures to the similar labeled measures based on which, pseudo-labels for the unlabeled data in each measure are inferred.  The mapping between the labeled and unlabeled measures based on the measure of measures OT is formulated as follows: 

Given an  image $z_i \in \mathbb{R}^{m\times n}$ from the either labeled or  unlabeled dataset, the CNN acts as a function $f (w, z_i) :\mathbb{R}^{m\times n} \rightarrow \mathbb{R}^c$ with the parameters $w$ that maps  $z_i$ to a c-dimensional representation, where c is number of the classes. Assume that $X=\{x_1,...,x_m\}$ and $X'=\{x'_1,...,x'_m\}$ are the sets of  c-dimensional outputs represented by the CNN for the labeled and unlabeled images, respectively. Let $\mathcal{P}_i= {1}/{n_i} \sum_{j=1}^{n_i} \delta_{x_{j}}$ denote a discrete measure constructed by the labeled data belonging to the $i$-th class, where $\delta_{x_j}$ is a Dirac unit mass on $x_j$ and $n_i$ is number of the data within the $i$-th class. Thus, all the labeled data construct a measure of measures $\mathcal{M}=\sum_{i=1}^c  \alpha_i \delta_{\mathcal{P}_i}$, where $\alpha_i={n_i}/{m}$ represents amount of the mass in the measure $\mathcal{P}_i$ and $\delta_{\mathcal{P}_i}$ is a Dirac unit mass on the measure $\mathcal{P}_i$. Similarly unlabeled data construct a measure of measures $\mathcal{M'}=  \sum_{j=1}^c  \beta_j \delta_{\mathcal{Q}_j}$ in that each measure $ \mathcal{Q}_i$, is created by the unlabeled data belonging to the unknown but the same class, where $\beta_j={n'_j}/{m}$ is amount of the mass in the measure ${\mathcal{Q}_j}$ and $\delta_{\mathcal{Q}_i}$ is a Dirac unit mass on ${\mathcal{Q}_j}$.

The goal of our SSL method is to use the OT to  find a coupling between the measures in $\mathcal{M'}$ and $\mathcal{M}$ that is optimal in the sense that it has a minimal expected transportation cost. This is because the transportation cost between two empirical measures which come from the same distribution (data from the same class) is expected to be less than the case where these measures come from two different distributions (data from different classes).  Thus,  we design an OT cost function defined in Eq. (3) to obtain an optimal coupling between measures in $\mathcal{M'}$ and $\mathcal{M}$ based on which the labels of data in the unlabeled measures are inferred:
\begin{equation}
f( \alpha, \beta, X)=\underset{T \in \mathcal{T}( \alpha, \beta)}{\mathrm{min}} \big \langle T, X\big \rangle-\lambda E(T),
\end{equation}
where $T$ is the optimal coupling matrix in which $T(i,j)$ indicates amount of the mass that should be moved from $\mathcal{Q}_i$ to $\mathcal{P}_j$ to provide an OT plan between $\mathcal{M'}$ and $\mathcal{M}$. Thus, if highest amount of the mass from $\mathcal{Q}_i$ is transported to $\mathcal{P}_k$ (i.e., $\mathcal{Q}_i$ is mapped to $\mathcal{P}_k$); the data belonging to the measure $\mathcal{Q}_i$ are annotated by $k$ which is the label of the measure $\mathcal{P}_k$. Variable $X$ is the pairwise similarity matrix between measures within
$\mathcal{M}$ and $\mathcal{M'}$ in which $X(i,j)=W_k(\mathcal{Q}_i, \mathcal{P}_j)$ which is the Wasserstein distance between two clouds of data points $\mathcal{Q}_i$ and $\mathcal{P}_j$. Note that the ground metric used for computing $W_k(\mathcal{Q}_i, \mathcal{P}_j)$ is the Euclidean distance.  Moreover, $\big \langle T,M \big \rangle$ denotes the Frobenius dot-product
between $T$ and $X$ matrices, and $\mathcal{T}$ is transportation polytope defined as follows:  $\mathcal{T}(\alpha,\beta)=\{T  \in \mathbb{R}^{c\times c}| T^\top \textbf{1}_c=\beta, T \textbf{1}_c=\alpha\}$ where $\textbf{1}_c$ is a c-dimensional vector with all elements equal to one. Finally, $E(T)$ is entropy of the  optimal coupling matrix $T$ which is used for regularizing the OT, and $\lambda$ is  a hyperparameter that balances between two terms in Eq. (3). The optimal coupling solution for the regularized OT defined in Eq. (3) is obtained by an iterative algorithm relied on  Sinkhorn algorithm (Appendix D). 
\section{Wasserstein Barycenters for Exploring Unlabeled Measures}
In Sec. 4, we represented the pool of unlabeled data as a measure of measures $\mathcal{M'}= \sum_{j=1}^c  \beta_j \delta_{\mathcal{Q}_j}$ 
in which each measure is constructed by data that belong to the same class.
However, label of the unlabeled data is unknown to allow us to identify these unlabeled measures. Moreover, CNN as a classifier trained on a limited amount of the labeled data simply miss-classifies these unlabeled data. In such a case,   there is little option other than to use unsupervised methods, such as the clustering to explore the unlabeled data belonging to the same class. This is because in structural assumption based on the clustering, it is assumed that the data within the same cluster are more likely to share the same label. Here, we leverage the Wasserstein metric to explore these unknown measures underlying the unlabeled data. Specifically, we relate the clustering algorithm to the problem of exploring  Wasserstein barycenter of the unlabeled data.

Wasserstein barycenter was initially introduced by  \cite{agueh2011barycenters}. Given probability measures $\mathcal{R}_1, ..., \mathcal{R}_l \in S_2(\theta)$ for $l \geq 1$, their Wasserstein barycenter $\tilde{\mathcal{R}}_{l,\mu}$ is defined as follows:
\begin{equation}
\tilde{\mathcal{R}}_{l,\mu}=\underset{\mathcal{R} \in S_2(\theta)}{\mathrm{argmin}}\sum_{i=1}^l \mu_i W_2^2(\mathcal{R},\mathcal{R}_i), 
\end{equation}
where $\mu_i$ is the weight associated with
$\mathcal{R}_i$. In the case where $\mathcal{R}_1,..., \mathcal{R}_l$ are discrete measures with finite number of elements and the weights in $\mu$ are uniform, it
is shown by \cite{anderes2016discrete} that the problem of exploring Wasserstein barycenter $\tilde{\mathcal{R}}_{l,\mu}$ on the space of $S_2(\theta)$
in (4) is recast to search only on $\mathcal{O}_r(\theta)$ denoting as a set of probability measures
with at most $r$ support points in $\theta$, where $r= \sum_{i=1}^l e_i-l+1$ and $e_i$ is the number of
elements in $\mathcal{R}_i$ for all $1 \leq i \leq l$. Moreover, an efficient algorithm for exploring local solutions of the Wasserstein barycenter problem over $\mathcal{O}_r(\theta)$ for some $r \geq 1$ has been studied by  \cite{cuturi2014fast}.

Beside, the popular K-means clustering can be considered as solving an optimization problem that comes up in the quantization problem, a simple but very practical connection \cite{pollard1982quantization,graf2007foundations}. The connection is as follows: Given $m$ unlabeled data $x'_1, . . . , x'_m \in \theta$. Suppose
that these data are related to at most $k$ clusters where $k \geq 1$ is a given number. The K-means problem finds
the set $Z$ containing at most $k$ atoms $\theta_1, . . . , \theta_k \in \theta$ that minimizes: $
\inf_{Z:|Z| \leq k}{\frac{1}{m} \sum_{i=1}^m d^2(x'_i,Z)}$.

Let $\mathcal{Q}=\frac{1}{m}\sum_{i=1}^m \delta_{x'_i}$ be a  measure created by data $x'_1, . . . , x'_m$. Then, $
\inf_{Z:|Z| \leq k}{\frac{1}{m} \sum_{i=1}^m d^2(x'_i,Z)}$ is equivalent to explore a discrete measure $\mathcal{H}$ including finite number of
support points and minimizing the following objective: $
\inf_{\mathcal{H} \in \mathcal{O}_k(\theta)}{ \sum_{i=1}^m W_2^2(\mathcal{H},\mathcal{Q})}.$ This problem can also be thought of as a Wasserstein barycenter problem when $l = 1$.  From this prospective, as denoted by \cite{cuturi2014fast}, the algorithm for finding the Wasserstein barycenters is an alternative for the popular Loyd’s algorithm to find local minimum of the K-means objective. Thus, we adopt the algorithm introduced in \cite{cuturi2014fast} used for computing the Wasserstein barycenters of empirical probability measures to explore the clusters underlying the unlabeled data (Appendix B). 

\section{ Training CNN in SSL Fashion with pseudo-labels}
Our SSL method finally leverages the unlabeled image data annotated by pseudo-labels obtained from the OT in conjunction with the supervision signals of the initial labeled image data to train the CNN classifier. Thus, we use the generic cross entropy as our discriminative
loss function to train the parameters of our CNN as follows: Let $\mathcal{X}_l$ be all of the labeled training data annotated by true labels $\mathcal{Y}$, and $\mathcal{X}_u$ be the unlabeled training data annotated by pseudo-labels  $\mathcal{Y'}$, then the total loss function $\mathcal{L}(.)$, used to train our CNN in an SSL fashion is as follows:   
\begin{equation}
\mathcal{L}(w,\mathcal{X}_l,\mathcal{X}_u,\mathcal{Y},\mathcal{Y'})=\mathcal{L}_c(w,\mathcal{X}_l,\mathcal{Y})+\alpha \mathcal{L}_c(w,\mathcal{X}_u,\mathcal{Y'}), 
\end{equation}
where $w$ is  parameters of the CNN, and  $\mathcal{L}_c(.)$ denotes cross entropy loss function, and $\alpha$ is a hyperparameter that balances between two losses obtained from the labeled and unlabeled data. For training, we initially train the CNN using the labeled data as a warm up step, and then use OT to provide pseudo-labels for the unlabeled data to train the CNN in conjunction with the initial labeled data for the next epochs. Specifically, after training the CNN using the labeled data, in each epoch, we select the same amount of initial labeled data from the pool of unlabeled data and then use OT to compute their pseudo-labels; then, we train  the CNN in a mini-batch mode. Our overall SSL method is described in Algorithm 2 (Appendix C).
\section{Experiments and Set-up} 
For evaluating our SSL technique and comparing it with the other SSL algorithms, we follow the concrete suggestions and criteria  which are provided in \cite{oliver2018realistic}. Some of these recommendations are as follows: 1) we use a common CNN architecture and training procedure to conduct a comparative analysis,  because differences in CNN
architecture or even implementation details can influence the results. 2) We report the performance of a fully-supervised case as a baseline because the goal of SSL is to greatly outperform the fully-supervised settings. 3) We change the amount of labeled and unlabeled data when reporting the performance of our SSL algorithm because an ideal SSL method should remain efficient even with the small amount of labeled and additional unlabeled data. 4) We also perform an analysis on  realistic small validation sets. This is because, in real-world applications, the large validation set is instead used as the training, therefore, an SSL algorithm which needs heavy tuning on a per-task  or per-model basis to perform well would not be applicable if the validation sets are realistically small (This analysis is done in Appendix F).


For the first criterion, we have used the 'WRN-28-2' model  (i.e., ResNet with depth 28 and width 2) \cite{zagoruyko2016wide}, including batch normalization \cite{ioffe2015batch} and leaky ReLU nonlinearities \cite{maas2013rectifier}. We conducted our experiments on the widely used  CIFAR-10 \cite{krizhevsky2009learning}, and SVHN \cite{netzer2011reading} datasets. Note that in our experiments, we tackle the general SSL challenge where the labeled and unlabeled data come from the same underlying distribution, and a given unlabeled data belongs to one of the classes in the labeled set and therefor, there is no class distribution mismatch. Moreover, for each of these datasets, we split the training set into two different sets of labeled and unlabeled data.  For training,  we use the well-known  Adam optimizer \cite{kingma2014adam}  with the default hyperparameters values and a learning rate of $3\times 10^{-3}$ in our experiments, and all the experiments have been done on a NVIDIA TITAN X GPU. The batch size in our experiments is set to 100. We have not used any form of early stopping; however, we have consistently monitored the performance of the validation set and reported test error at the point of lowest validation error.  The stopping criteria for the Sinkhorn algorithm is either maxIter $=$ 10,000 or tolerance $=10^{-8}$,  where maxIter is  the maximum number of iterations and tolerance is a threshold for the integrated stopping criterion based on the marginal differences. In  experiments, we  followed the data augmentation and standard data normalization used in \cite{oliver2018realistic}.  Specifically, for SVHN, we
converted pixel intensity values of the images to floating point values in the range of [-1, 1]. For the data augmentation, we only applied random translation by up to 2 pixels. We used the standard training and validation split, with 65,932 images for the training set and 7,325 for the validation set.
For CIFAR-10, we applied global contrast normalization.  The data augmentation on
CIFAR-10 are random translation by up to 2 pixels, random horizontal flipping, and Gaussian
input noise with standard deviation 0.15. We used the standard training and validation split, with 45,000
images for the training set and 5,000 images for the validation set.
\subsection{Fully Supervised Baseline and Deep SSL Methods}
\begin{table*}
    \centering
    \scalebox{.64}{\begin{tabular}{c c c c c c c c c c c}
     Methods & \# Labels & Supervised & ROT & Soft-ROT & $\pi$ Model & Mean Teacher & VAT  & VAT + EntMin & Pseudo-Label  \\
         \hline
      CIFAR-10 & $4000$ & ${20.89} (\pm {0.47})$ & ${6.06} (\pm {0.12})$ & ${6.82} (\pm {0.17})$ & ${16.37} (\pm {0.63})$ & ${15.87} (\pm {0.28})$ & ${13.86} (\pm {0.27})$ & ${13.13} (\pm {0.39})$ & ${17.78} (\pm {0.57})$\\
       SVHN  & $1000$ & ${13.11} (\pm {0.53})$ & ${3.11} (\pm {0.45})$ & ${3.51} (\pm {0.49})$  & ${7.19} (\pm {0.27})$ & ${5.65} (\pm {0.47})$ & ${5.63} (\pm {0.20})$ & ${5.35} (\pm {0.19})$ & ${7.62} (\pm {0.29})$
    \end{tabular}}
    \caption{ Comparing  deep SSL models using test error rate on SVHN, and CIFAR-10.
}
    \label{tabel1}
    \vspace{-6mm}
\end{table*}
Here, we consider the second criterion for evaluation of our SSL method.  The purpose of SSL is mainly to achieve a better performance when it uses the unlabeled data than the case where using the labeled data alone.  To ensure that our SSL model benefits from the unlabeled data during the training, we report the error rate of the WRN model for both cases where we only use the labeled data (i.e., Supervised in Table. \ref{tabel1}), and the case where we leverage the unlabeled data by using the  OT technique during the training (i.e., ROT in Table. \ref{tabel1}). Moreover, we have reported the performance of other SSL algorithms in  Table. \ref{tabel1} which also leverage the unlabeled data during the training. All of the compared SSL methods use the common CNN model (i.e., 'WRN-28-2') and training procedure as suggested in the first criterion for the realistic evaluation of SSL models. The result of all SSL methods reported in Table. \ref{tabel1} is the test error at the point of lowest validation error for tuning their hyperparameters. For a fair evaluation with other SSL algorithms, we selected 4,000 samples of the training set as the labeled data and the remaining as the unlabeled data for the CIFAR-10 dataset, and we chose 1,000 samples of the training set as the labeled data and the rest as the unlabeled data for the SVHN dataset. We ran our SSL algorithm over five times with different random splits of  labeled and
unlabeled sets for each dataset, and we reported the mean and standard deviation of the test error rate in  Table. \ref{tabel1}. The results in  Table. \ref{tabel1} indicates that on both CIFAR-10 and SVHN, the gap between the fully-supervised
baseline and ROT is bigger  than this gap for the other SSL methods. This indicates the potential of our model for leveraging the unlabeled data in comparison to other methods that also use the unlabeled data to improve the classification performance of a CNN model in SSL fashion. Moreover, we trained our baseline WRN on the entire training set of CIFAR-10 and SVHN and the test error over five runs are ${4.23} (\pm {0.18})$ and ${2.56} (\pm {0.04})$, respectively.

Besides the particular manner in which we choose the one particular pseudo-label, we also use "soft pseudo-labels". Essentially, instead of having the one-hot target in the usual classification loss (i.e., cross-entropy), we can have the row of the transport plan corresponding to the unlabeled data points as the target. We used the soft pseudo-labels produced by OT to train the CNN. The comparison of results in Table. \ref{tabel1}  show that one-hot targets used in ROT outperforms the soft pseudo-labels used in ROT. Why this is happening can be supported by SSL methods based on the entropy minimization criterion. This set of methods force the model to produce confident predictions (i.e., low entropy for output of the model). Similarly here, once we use one-hot targets, we encourage the network to produce more confident predictions than when using soft-pseudo labels.
\begin{table*}
    \centering
    \scalebox{.7}{\begin{tabular}{c c c c c c}
     Methods & \# Labels & ROT & Soft-ROT & S-M-GNN & S-S-GNN \\
         \hline
      CIFAR-10 & $4000$ & ${6.06} (\pm {0.12})$ & ${6.82} (\pm {0.17})$& ${13.95} (\pm {0.53})$ & ${18.63} (\pm {0.32})$ \\
       SVHN  & $1000$ & ${3.11} (\pm {0.45})$ & ${3.51} (\pm {0.49})$& ${7.91} (\pm {0.34})$ & ${11.89} (\pm {0.48})$
    \end{tabular}}
    \caption{Comparing test error over five runs between  ROT and S-S-GNN and S-M-GNN baselines.  
}
    \label{tab1}
    \vspace{-3mm}
\end{table*}
 \begin{figure}[t]
\centering
\subfigure[\label{fig2a} CIFAR-10]{\includegraphics[scale=0.46]{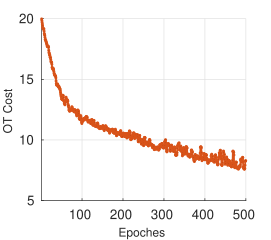}}
\subfigure[\label{fig2b}SVHN]{\includegraphics[scale=0.46]{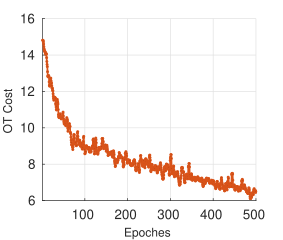}}
\subfigure[\label{fig2c}CIFAR-10]{\includegraphics[scale=0.46]{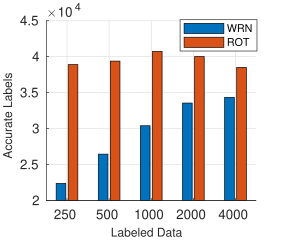}}
\subfigure[\label{fig2d}SVHN]{\includegraphics[scale=0.46]{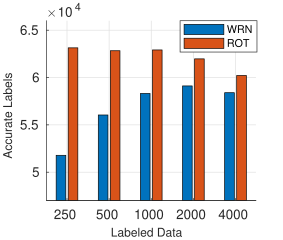}}
\label{cam3}
\caption{ a) and b) are the OT cost between the labeled and unlabeled measures during the training, c) and d) indicate the number of accurate predicted labels for the remaining training unlabeled data.}
\vspace{-5mm}
\end{figure}
\subsection{OT Baselines for SSL }
In this section, we compare ROT which is based on the measure of measure OT with two other baselines. Both the baselines assign pseudo-labels for the unlabeled samples based on the greedy nearest neighbor (GNN) search. The first baseline is sample to sample (S-S-GNN) case, where pseudo-labels for the unlabeled data are obtained by GNN on the outputs of softmax layer. Specifically, for each of the unlabeled sample, we annotate it with the label of the closest labeled sample in the training set. The second baseline is sample to measure (S-M-GNN) case where, pseudo-labels of the unlabeled samples are obtained based on the GNN between the unlabeled samples and the probability measures constructed by initial labeled data in the training set. When transporting from a Dirac to a probability measure, the OT problem (regularized or not) has a closed form. Essentially, there is only one admissible coupling. Thus, in such a case,  the Wasserstein distance between a sample to a probability measure  is simply computed as follows: Given an unlabeled Dirac $\delta_{x'_i}$ and a labeled measure $\mathcal{P}_j=\sum_{i=1}^m a_i\delta_{x_i}$, then $W_k(x'_i,\mathcal{P}_j)= \sum_{k=1}^m a_k||x'_i-x_k||^k$. 

The comparison of results between ROT, and these baselines on the  SVHN and CIFAR-10 in Table. \ref{tab1} shows the benefit of measure of measure OT for training a CNN in an SSL manner.

\subsection{Contribution of Optimal Transport to Deep SSL}
 Instead of using the CNN as a classifier to produce pseudo-labels for the unlabeled data, we used the Wasserstein barycenters to cluster the unlabeled data. This allowed us to explore the unlabeled measures that we could then match them with the labeled measures for pseudo-labeling. This was because the CNN,  as a classifier trained on a limited amount of the labeled data, simply miss-classifies the unlabeled data. To compare these two different strategies for producing the pseudo-labels to train the CNN classifier in an SSL fashion,  we experimentally show  how the clustering-based method (i.e., ROT) can have a greater positive influence on the training of our CNN classifier. We report the number of  pseudo-labels which are accurately predicted by ROT. This result allows us to know the level of accuracy of the pseudo-label obtained for the unlabeled data, which the CNN can then benefit from during the training. We  also report these results with that of predicted labels achieved by the baseline CNN classifier (i.e., WRN) on the unlabeled training data. This comparison also allows us to know whether or not the CNN classifier can benefit from our strategy for providing pseudo-labels during the training, because, otherwise, the WRN can simply use its own predicted labels on unlabeled training data over the course of training.  To indicate the efficiency of our method during the training of the CNN, we changed the number of initial labeled data in the training set and reported the number of accurately predicted pseudo-labels by the baseline WRN, and ROT on the remaining unlabeled training data. Fig.  \ref{fig2c} and Fig.  \ref{fig2d} show that, for both CIFAR and SVHN datasets, the labels predicted by ROT on the unlabeled training data are more accurate than the WRN, which means that the entire CNN network can better benefit from the ROT strategy than the case where it is trained solely by its own predicted labels. Moreover, we monitored the trend of transportation cost between the labeled and unlabeled measures obtained by Eq. 3 during the training.  Fig.  \ref{fig2a} and Fig.  \ref{fig2b}  show that the transportation cost is reduced as the images fed into the CNN are represented by a better feature set during the training. 
 \begin{figure}[t]
\centering
\subfigure[\label{fig2a}CIFAR-10]{\includegraphics[scale=0.43]{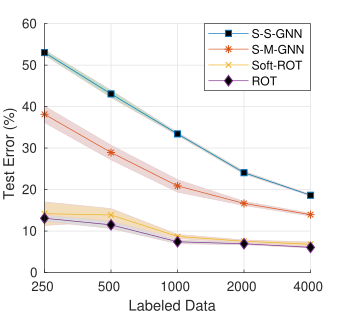}}
\subfigure[\label{fig2b}SVHN]{\includegraphics[scale=0.43]{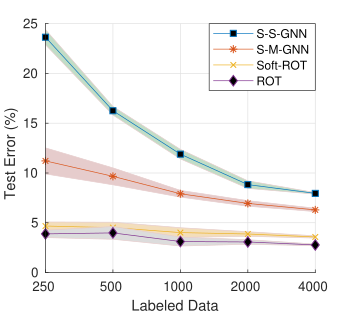}}
\subfigure[\label{fig2c}SVHN]{\includegraphics[scale=0.42]{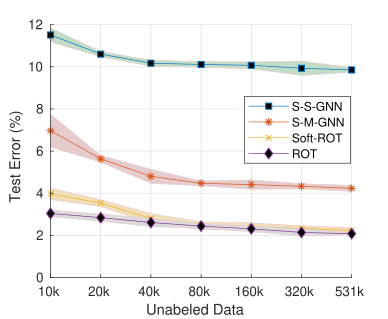}}
\label{cam3}
\caption{Test error of the ROT algorithm with varying amount of labeled and unlabeled data.}
\vspace{-5mm}
\end{figure}
\subsection{Varying the Amount of Labeled and Unlabeled Data}
In Table. \ref{tab1}, we evaluated ROT for the case where we only use 4,000  and 1,000  initial labeled data for the CIFAR-10 and SVHN, respectively. However, here, we explore that how varying the amount of initial labeled data  decreases the performance of ROT  in the very limited label regime, and also at which point our SSL method can recover the performance of training when using all of the labeled data in the dataset. To do this evaluation, we gradually increase the number of labeled data during the training and report the performance of our SSL method on the testing set. In this experiment, we ran our SSL method over five times with different random splits of labeled and unlabeled sets for each dataset, and reported the mean and standard deviation of the error rate in Fig. \ref{fig2a} and Fig. \ref{fig2b}. The results show that the performance of ROT tends to converge as the number of labels increases. 

Another possibility for evaluating the performance of our SSL method is to change the number of unlabeled data during the training. However, using the CIFAR-10 and SVHN
datasets in isolation puts an upper limit on the  amount of available unlabeled data. Fortunately, in contrast to CIFAR-10, SVHN has been distributed with the “SVHN-extra” dataset, which includes 531,131 additional digit images and  has also been previously used as unlabeled data for evaluation of different SSL methods in  \cite{oliver2018realistic}. These additional data come from the same distribution as SVHN does, which allows us to use them in our SSL framework. Fig. \ref{fig2c} shows the trend of test error for our SSL algorithm on SVHN
with 1,000 labels and changing amounts of unlabeled
images from SVHN-extra dataset. The results shows that, increasing the amount of unlabeled data improves the performance of our SSL method, but this improvement is not significant when we provide 40k unlabeled data. 



\section{Conclusion}

  We proposed a new SSL method based on the optimal transportation technique in which unlabeled data masses are transported to a set of labeled data masses, each of which is constructed by data belonging to the same class. In this method, we found a mapping between the labeled and unlabeled masses which was used to infer pseudo-labels for the unlabeled data so that we could use them to train our CNN model. Finally, we experimentally evaluated our SSL method to indicate its potential and effectiveness for leveraging the unlabeled data when labels are limited during the training.

\bibliography{iclr2020_conference}
\bibliographystyle{iclr2020_conference}
\newpage
\appendix

\section{Background and Definitions}
\textbf{Discrete Optimal Transport:} For any $r \geq 1 $, let the probability simplex be denoted
by $\Delta_r =\{v \in \mathbb{R}^r : v_i\geq 0, \sum_{i=1}^r v_i=1\}$, and also assume that   ${U}=\{u_1,...,u_n\}$ and $V=\{v_1,...,v_m\}$ are two sets of data points in $\mathbb{R}^d$ such that $\mathcal{U}=\sum_{i=1}^n a_i\delta_{u_{i}}$ and $\mathcal{V}=\sum_{i=1}^m b_i\delta_{v_{i}}$; the Wasserstein distance  $W_k(\mathcal{U}, \mathcal{V})$ between two discrete measures $\mathcal{U}$ and $\mathcal{V}$ is the $k$-th root of the optimum of a network flow problem known as the transportation problem \cite{bertsimas1997introduction}. Note that $\delta_{u_i}$ is the Dirac unit mass located on point $u_i$, $a$ and $b$ are the weighting vectors which belong to the probability  simplex $\Delta_n$ and $\Delta_{m}$, respectively. The transportation problem depends on the two following components: 1)  matrix $M \in \mathbb{R}_+^{n\times m} $ which encodes the geometry of the data points by measuring the pairwise distance between elements in $U$ and $V$ increased to the power $k$,  2) the transportation polytope $P(a,b) \in \mathbb{R}_+^{n\times m}$ which acts as a feasible set, characterized as a set of
$n \times m$ non-negative matrices such that their row and column
marginals are  $a$ and $b$, respectively. This means that the transportation plan should satisfy the marginal
constraints. In other words, let $\textbf{1}_m$ be an $m$-dimensional vector with all elements equal to one, then the transportation polytope is represented as follows:  $P(a,b)=\{T  \in \mathbb{R}_+^{n\times m}| T^\top \textbf{1}_n=b, T \textbf{1}_m=a\}$. Essentially, each element $T(i,j)$ indicates the amount of  mass which is transported from $i$ to $j$. Note that in the transportation problem, the matrix $M$ is also considered as a cost parameter such that $M(i,j)= D^k(u_i,v_j)$ where $D(.)$ is the Euclidean distance.  

Let $\big \langle T,M \big \rangle$  denote the Frobenius dot-product
between $T$ and $M$ matrices. Then the discrete Wasserstein distance $W_k(\mathcal{U}, \mathcal{V})$ 
is formulated by an optimum of a parametric linear program $g(.)$ on a cost matrix $M$, and $n \times m$ number of variables parameterized by the marginals
$a$ and $b$ as follows:
\begin{equation}
W_k(\mathcal{U}, \mathcal{V})=g(a, b, M)=\underset{T \in P(a, b)}{\mathrm{min}} \big \langle T, M \big \rangle.
\end{equation}

The  Wasserstein distance in (6) is a Linear Program (LP) and a subgradient of its solution can be calculated using Lagrange duality. The dual LP of (6) is formulated as follows:
\begin{equation}
\textbf{d}(a, b, M)=\underset{(\alpha,\beta) \in C_M}{\mathrm{max}} \alpha^\top a+\beta^\top b,
\end{equation}
where the polyhedron $C_M$ of dual variables is as follows:
\begin{equation}
C_M=\{(\alpha,\beta)\in \mathbb{R}_+^{m+n}|\alpha_i+\beta_j \leq M(i,j)\}.
\end{equation}
Considering LP duality, the following  equality is established $\textbf{d}(a, b, M) = \textbf{p}(a, b, M)$ \cite{bertsimas1997introduction}. Computing the exact Wasserstein distance in (6) is time consuming. To alleviate this problem,  \cite{cuturi2013sinkhorn} has introduced an interesting method that regularizes (6) using the entropy of the solution matrix $H(T)$, (i.e., ${\mathrm{min}} \big \langle T, M \big \rangle +\gamma H(T)$). It has been shown that if $T'_{\gamma}$ is the solution of the regularized version of (6) and $\alpha'_{\gamma}$  is its dual solution in (7), then $\exists! u\in \mathbb{R}_+^n$, $v\in \mathbb{R}_+^m$ such that the solution matrix is  $T'_{\gamma}=\text{diag}(u)K\text{diag}(v)$ and $\alpha'_\gamma=-\log(u)/\gamma+(\log(u)^\top\textbf{1}_n)/({\gamma n})) \textbf{1}_n$ where, $K=exp(-M/\gamma)$. The vectors $u$ and $v$ are updated iteratively between step 1 and 2 by using the well-known Sinkhorn algorithm as follows: step $1) u=a/{Kv}$ and step $2) v=b/{K^{\top}u}$, where$/$ denotes element-wise division operator \cite{cuturi2013sinkhorn}.
\section{Wasserstein barycenter of the unlabeled Data}
Given an  image $x_n \in \mathbb{R}^{m\times n}$ from the either labeled or the unlabeled set, the CNN acts as a function $f_n :\mathbb{R}^{m\times n} \rightarrow \mathbb{R}^c$ with the parameters $\theta_n$ that maps  $x_n$ to a c-dimensional representation, where c is the number of classes. Assume that $X_u=\{x'_1,...,x'_n\}$ is the set of  CNN outputs extracted from the unlabeled data. As noted in \cite{cuturi2014fast}, the Wasserstein barycenter of the unlabeled set $X_u$ is equivalent to Lloyd's algorithm, where the maximization step (i.e.,  the assignment of the weight of each data point to its closest centroid) is equivalent to
the computation of $\bm{\alpha}'$ in dual form, while the expectation step (i.e., the re-centering step) is equivalent to  the update for centers $Y$ using
the optimal transport, which in this case is equivalent to the trivial transportation plan that assigns the weight (divided by $n$) of each unlabeled data in $X_u$ to its closest neighbor in centers $Y$. Algorithm 1 shows the Wasserstein barycenter of the unlabeled data for clustering.
\begin{algorithm}[H]
\caption{: Wasserstein barycenter of the unlabeled Data }
\textbf{input:} $X_u \in \mathbb{R}^{c\times n}, b \in \Delta_n$
\begin{algorithmic}[1]
\label{algo1}
\STATE {\textbf{initialize}:  $Y \in \mathbb{R}^{c\times k}$ and $a \in \theta$} 
\WHILE{$Y$ and $a$ have not converged} 
\STATE \textbf{Maximization Step:}
\STATE set $\hat{a} = \tilde a= \nicefrac {\textbf{1}_n}{n}$
 \WHILE{not converged}
\STATE $\beta=(t+1)/2$, $a\leftarrow{(1-\beta^{-1})\hat{a}+\beta^{-1}\tilde a}$
\STATE  $\alpha \leftarrow$ $\bm{\alpha}'$ dual optimal form $\textbf{d}(a,b, M_{X_u Y})$
\STATE $\tilde a \leftarrow \tilde a \circ e^{-t_0\beta \alpha}; \tilde a \leftarrow \nicefrac{\tilde a}{\tilde a^ \top} \textbf{1}_n$
\STATE $\hat{a}  \leftarrow (1-\beta^{-1})\hat{a}+\beta^{-1}\tilde a$, $t \leftarrow t+1$
\ENDWHILE
\STATE $a\leftarrow \hat{a} $
\STATE  \textbf{Expectation Step:}
\STATE $T'\leftarrow$ optimal coupling of $\textbf{p}(a,b,M_{X_u Y})$
\STATE $Y\leftarrow (1-\theta) Y+\theta (XT^{'\top})\text{diag}(a^{-1})$, $\theta \in [0,1]$
\ENDWHILE
\end{algorithmic}
\end{algorithm}
\section{Matching Distributions via OT for Semi-Supervised Learning}

\begin{algorithm}[H]
\caption{: Matching Distributions via  OT for SSL}
\textbf{input:} labeled data: $Z_l=\{z_l, y_l\}_{l=1}^n$, unlabeled data: $Z_u=\{z'_u\}_{u=1}^m$, balancing coefficients: $\alpha$, $\lambda$, learning rate: $\beta$, batch size: $b$,  distance matrix: $X$,
\begin{algorithmic}[1]
\label{algo1}
\STATE train CNN parameters initially using the labeled data,
\REPEAT
\STATE  $X_l=\{x_l\}_{l=1}^n$, $X_u=\{x'_u\}_{u=1}^n$: Softmax layer outputs on $Z_l$ and $Z_u$,
\STATE $ \{\mathcal{Q}_1, ..., \mathcal{Q}_c\} \gets \text{cluster on  $X_u$ using Algorithm. 1}$,
\STATE $ \{\mathcal{P}_1, ..., \mathcal{P}_c\} \gets \text{labeled data grouped to $c$ classes}$,
\STATE compute $\alpha$, $\beta$ based on  amount of the mass in measures  $\mathcal{Q}$ and $\mathcal{P}$,
\FOR{ each $\mathcal{Q}_i$ and $\mathcal{P}_j$ }

\STATE $X (i,j) \gets W_2(\mathcal{Q}_i,\mathcal{P}_j)$,
 
 \ENDFOR
 
\STATE $T \gets$ optimal coupling of $\textbf{p}(\alpha,\beta ,X)$,
 
 \STATE {$\{y'_u\}_{u=1}^n \gets$ pseudo-label data in each cluster $\mathcal{Q}_i$ with the highest amount of mass transport toward the labeled measure} (i.e., $\operatorname*{argmax} T(i,:)$),
\REPEAT
\STATE choose a mini-batch:$\{x_i\}_{i=1}^b \subset X_u \cup X_l$,
\STATE $ w \gets w -\beta \nabla_{w} [ \mathcal{L}(w,x,x', y,y')]$,  using  Eq. (5),
\UNTIL for an epoch
\UNTIL  a fixed number of epochs
\end{algorithmic}
\end{algorithm}
\section{Relaxing Optimization via Entropic Regularization}
The regular OT problem defined in (6) can be solved by an effective linear programming method  in the order of $\mathcal{O}(n^3 log (n))$ time complexity, where  $n$ is  number of the points in each probability measures.  Cuturi \cite{cuturi2013sinkhorn} has introduced  an interesting approach  which relaxes the OT problem by adding a strong convex regularizer to the OT cost function to reduce the time complexity to $\mathcal{O}(n^2)$. Specifically, this approach asks for  a solution $T'$ with more entropy, instead of computing the exact Wasserstein distance. In other words, the regularized OT
distances can interpolate the solution, depending
on the regularization strength $\gamma$, between exact OT ($\gamma=0$ ), and Maximum Mean Discrepancy, MMD, ($\gamma=\infty$). In this work, we use the regularized OT not only for the matter of time complexity, but also it has been shown that the  sample complexity of exact Wasserstein distance is $O(1/n^{1/d})$, while the regularized  Wasserstein distance depending on $\gamma$ value, is between $O(1/\sqrt{n})$ and $O(1/n^{1/d})$, where $d$ is  dimension of the samples \cite{pmlr-v89-genevay19a,peyre2019computational}. This means that the entropic regularization reduces the chance of over-fitting for our SSL model when it computes the Wasserstein distance between output of the CNN obtained from the labeled and unlabeled data. Hence, our OT problem  in the regularized form is recast as follows:
\begin{equation}
\tilde{W}_{\gamma}(\mathcal{M'}, \mathcal{M})=\underset{T' \in P'(a, \alpha)}{\mathrm{min}} \big \langle T', X\big \rangle-\gamma E(T'),
\end{equation}
where $\gamma$ is a hyperparameter that balances two terms in (9), and   $E(T')=-\sum_{ij}^{mn}T'_{ij}(log (T'_{ij}-1)$ is the entropy of the solution matrix $T'$. It has been shown that if $T'_{\gamma}$ is the solution of the optimization (9), then $\exists! u\in \mathbb{R}_+^n$, $v\in \mathbb{R}_+^m$ such that the solution matrix for (9) is  $T'_{\gamma}=\text{diag}(u)K\text{diag}(v)$ where, $K=exp(-X/\gamma)$ \cite{cuturi2013sinkhorn}. The vectors $u$ and $v$ are updated iteratively between step 1 and 2 by using the well-known Sinkhorn algorithm as follows: step $1) u=a/{Kv}$ and step $2) v=b/{K^{\top}u}$, where$/$ denotes element-wise division operator \cite{cuturi2013sinkhorn}.
\section{Existence of Optimal Coupling for Measure of Measures}

 It can be simply shown that there always exists an optimal coupling, $\pi \in \Pi(\mathcal{M}, \mathcal{M}')$, that achieves  infimum of  Eq. (2) in the paper. This is because the cost function $||x-y||$ in Eq. (1) is continuous, and based on  Theorem 4.1, the existence of an optimal coupling $\pi \in \Pi(\mathcal{R}, \mathcal{S})$ which obtains the infimum is guaranteed due to the tightness of $\Pi(\mathcal{R}, \mathcal{S})$. Furthermore, based on  Corollary 6.11, the term $  W_k(x,x')$ used in  Eq. (2) is a continuous function and $\Pi (\mathcal{M},\mathcal{M}')$ is  tight again, so the existence of an optimal coupling in $\Pi (\mathcal{M},\mathcal{M}')$ is also guaranteed.

\textbf{Theorem 4.1} in  Villani’s book \cite{villani2008optimal}:  

Let $L^1$ be the Lebesgue space of exponent 1, and  $(\mathcal{X} ,\mu)$
and $(\mathcal{Y},\nu)$ be two Polish probability spaces; let $a : \mathcal{X} \rightarrow \mathbb{R} \cup \{-\infty\}$
and $b : \mathcal{Y} \rightarrow \mathbb{R} \cup \{-\infty \}$ be two upper semi-continuous functions such
that $a \in L^1(\mu)$, $b \in L^1(\nu)$. Let $c : \mathcal{X} \times \mathcal{Y} \rightarrow \mathbb {R}\cup \{+\infty\}$ be a lower
semi-continuous cost function, such that $c(x,y) \geq a(x) + b(y)$ for all
$x, y$. Then there is a coupling of $(\mu,\nu)$ which minimizes the total cost $\mathbb{E}c(X,Y )$ among all possible couplings $({X},{Y})$.

\textbf{Lemma 1}: Let $\mathcal{X}$ and $\mathcal{Y}$ be
two Polish spaces. Let $\mathcal{R} \subset \mathcal{P}(\mathcal{X})$ and $\mathcal{S} \subset \mathcal{P}(\mathcal{Y})$  be tight subsets of
$\mathcal{P}(\mathcal{X})$ and $\mathcal{P}(\mathcal{Y})$ respectively. Then, the set $\Pi(\mathcal{R}, \mathcal{S})$ of all transference
plans whose marginals lie in $\mathcal{R}$ and $\mathcal{S}$ respectively, is itself tight in
$\mathcal{P}(\mathcal{X} \times \mathcal{Y})$.

\textit{Proof of Lemma:} Let $\mu \in \mathcal{R}, \nu \in \mathcal{S}$, and $\pi \in \Pi (\mu,\nu)$. By assuming that, for any $\epsilon > 0$ there is a compact set $K_{\epsilon} \subset \mathcal{X}$ , independent of
the choice of $\mu$ in $\mathcal{R}$, such that $\mathcal{\mu}[\mathcal{X}\textbackslash K_{\epsilon}] \leq \epsilon$; and similarly there is
a compact set $L_{\epsilon} \subset \mathcal{Y}$, independent of the choice of $\nu$ in $S$, such that $\nu[\mathcal{Y}\textbackslash  L_{\epsilon}] \leq \epsilon$. Then, for any coupling $(X,Y)$ of $(\mu,\nu)$,
\begin{equation*}
    \mathbb{P}[(X,Y) \notin K_{\epsilon}\times L_{\epsilon}]\leq \mathbb{P}[X \notin K_{\epsilon}]+ \mathbb{P} [Y \notin L_{\epsilon}] \leq 2\epsilon.
\end{equation*}
The desired result follows because this bound is independent of the coupling, and $K_{\epsilon} \times L_{\epsilon}$ is compact in $\mathcal{X} \times \mathcal{Y}$.

\textbf{Lemma 2}: Let
$\mathcal{X}$ and $\mathcal{Y}$ be two Polish spaces, and $c : \mathcal{X} \times \mathcal{Y} \rightarrow \mathbb{R} \cup \{+ \infty\}$ a lower semi-continuous cost function. Let $h : \mathcal{X} \times \mathcal{Y} \rightarrow \mathbb{R} \cup \{-\infty \}$ be an upper
semi-continuous function such that $c \geq h$. Let $(\pi_k)_k \in N$ be a sequence of
probability measures on $\mathcal{X} \times \mathcal{Y}$, converging weakly to some $\pi \in \mathcal{P}(\mathcal{X}\times \mathcal{Y})$,
in such a way that $h \in L^1 (\pi_k)$, $h \in L^1(\pi)$, and
\begin{equation*}
   \int_{\mathcal{X}\times\mathcal{Y}} h d \pi_k \xrightarrow{k\rightarrow\infty}\int_{\mathcal{X}\times\mathcal{Y}} h d\pi.
\end{equation*}
Therefore, 
\begin{equation*}
   \int_{\mathcal{X}\times\mathcal{Y}} h d\pi \leq \lim_{k\rightarrow \infty}\inf \int_{\mathcal{X}\times\mathcal{Y}} c d \pi_k
\end{equation*}
In particular, if $c$ is non-negative, then $F : \pi \rightarrow
\int c d\pi$ is lower semi-continuous on $\mathcal{P}(\mathcal{X} \times \mathcal{Y})$, equipped with the topology of weak convergence.

\textit{Proof of Lemma:}  Replacing $c$ by $c - h$, we may assume that $c$ is
a non-negative lower semi-continuous function. Then $c$ can be written
as the point-wise limit of a non-decreasing family $(c_{\ell})_{\ell} \in \mathbb{N}$ of continuous real-valued functions. By monotone convergence,
\begin{equation*}
  \int c d \pi= \lim_{\ell\rightarrow \infty} \int c_{\ell} d \pi=\lim_{\ell \rightarrow \infty} \lim_{k\rightarrow \infty} \int c_{\ell} d \pi_k \leq \lim \inf_{k\rightarrow \infty} \int c d \pi_k.
\end{equation*}
\textbf{Prokhorov’s Theorem} \cite{billingsley2013convergence}: If $\mathcal{X}$ is a Polish space, then
a set $\mathcal{R} \subset \mathcal{P}(\mathcal{X})$ is pre-compact for the weak topology if and only if it is
tight, i.e. for any $\epsilon > 0$ there is a compact set $K_{\epsilon}$ such that $\mu[\mathcal{X} \textbackslash K_{\epsilon}] \leq \epsilon$
for all $\mu \in \mathcal{R}$.

\textbf{Proof of Theorem 4.1}: Since $\mathcal{X}$ is Polish, $\{\mu \}$ is tight in $\mathcal{P}(\mathcal{X})$; similarly,
$\{\nu\}$ is tight in $\mathcal{P}(\mathcal{Y})$. By using the Lemma 1, $\Pi(\mu,\nu)$ is tight in $\mathcal{P}(\mathcal{X} \times \mathcal{Y})$, and by using Prokhorov’s theorem, this set has a compact closure. By passing to
the limit in the equation for marginals, we see that $\Pi(\mu,\nu)$ is closed, so it is in fact compact. Then let $(\pi_k)_k\in \mathbb{N}$ be a sequence of probability measures on $\mathcal{X} \times \mathcal{Y}$,
such that $\int c d \pi_k$ converges to the infimum transport cost. Extracting
a sub-sequence if necessary, we may assume that $\pi_k$ converges to some $\pi \in \Pi(\mu,\nu)$. The function $h : (x,y) \rightarrow a(x) + b(y)$ lies in $L^1(\pi_k)$
and in $L^1(\pi)$, and $c \geq h$ by assumption; moreover, $\int
h d \pi_k = \int h d \pi = \int a d \mu +\int b d \nu$; so Lemma 2 implies:
\begin{equation*}
   \int c d \pi \leq \lim \inf_{k\rightarrow \infty}\int c d \pi_k.
\end{equation*}
Therefore, $\pi$ is minimizing.

Note that further details of the proof of Theorem 4.1 are also available in Villani's book \cite{villani2008optimal}.

\textbf{Corollary 6.11} in Villani’s book \cite{villani2008optimal}:

If $(\mathcal{X} ,d)$ is a Polish space, and $p \in [1,\infty)$, then $W_p$ is continuous on $\mathcal{P}_p(\mathcal{X})$. More explicitly, if $\mu_k$
(resp. $\nu_k$) converges to $\mu$ (resp. $\nu$) weakly in $\mathcal{P}_p(\mathcal{X})$ as $k \rightarrow \infty$, then
\begin{equation*}
    W_p(\mu_k,\nu_k) \rightarrow W_p(\mu,\nu).
\end{equation*}
\section{Hyperparameter Tuning on Realistically Small Validation Sets}

One of the interesting arguments presented in \cite{oliver2018realistic} for a standard evaluation of different SSL models is that it may not be feasible to perform model selection for an SSL challenge if the hyperparameters of the model are tuned on the realistically small validation sets. On the other hand, most of the SSL datasets in the literature are designed in such a way that the validation set, which is used for tuning the hyperparameters but not for parameters of the model, is much larger than the training set. For example, the standard SVHN dataset used in our work has about 7000 labeled data in the validation set. Hence, the validation set is seven times larger than the training set of the SSL methods which evaluate their performance by using only 1,000 labeled data during the training. However, this is not a practical choice for a real-world application. This is because,  this large validation set will be used as the training set instead of validation set for tuning the hyperparameters. Using small validation sets, however, causes an issue in that the evaluation metric, such as the accuracy for tuning the hyperparameters  will be  unstable and noisy across the different runs. 
 \begin{figure}[t]
\centering
\subfigure[\label{fig3a}]{\includegraphics[scale=0.43]{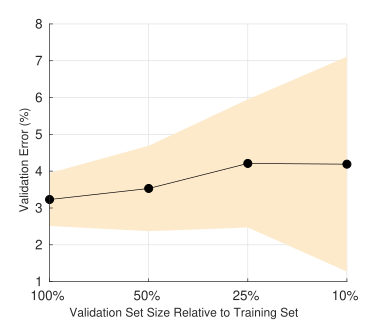}}
\subfigure[\label{fig3b}]{\includegraphics[scale=0.43]{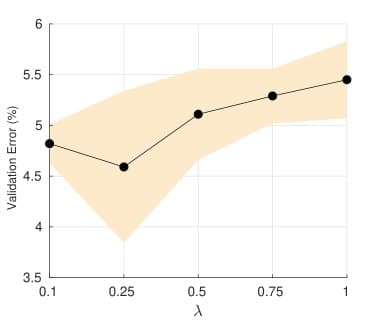}}
\subfigure[\label{fig3c}]{\includegraphics[scale=0.43]{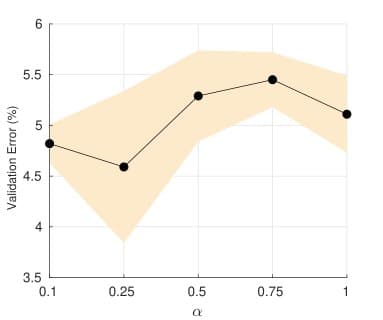}}
\label{cam3}
\caption{ The error of the ROT algorithm on SVHN validation set over five runs.}
\end{figure}
 \begin{figure}[t]
\centering
\subfigure[\label{fig4a}]{\includegraphics[scale=0.43]{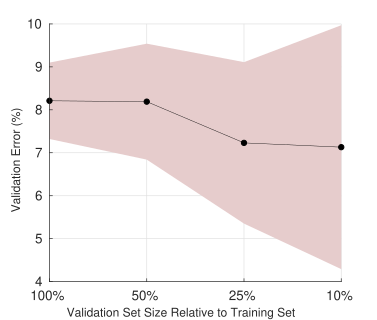}}
\subfigure[\label{fig4b}]{\includegraphics[scale=0.43]{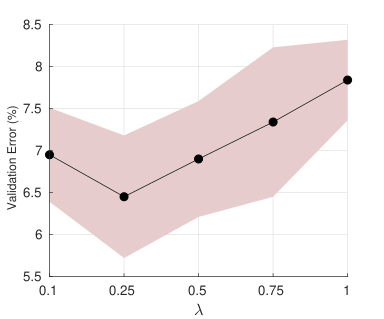}}
\subfigure[\label{fig4c}]{\includegraphics[scale=0.43]{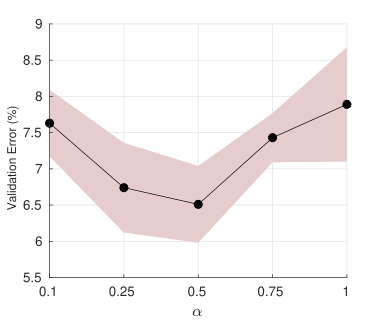}}
\label{cam3}
\caption{ The error of the ROT algorithm on CIFAR validation set over five runs.}
\end{figure}
Although  the fact that small validation
sets limit the ability for model selection has been discussed in \cite{chapelle2009semi}, the work presented in \cite{oliver2018realistic} has used  the Hoeffding inequality \cite{hoeffding1994probability} to directly  analyze the relationship between the size of validation set and the variance in estimation of a
model’s accuracy:
\begin{equation*}
  \mathbb{P}(|\overline{V}-\mathbb{E}(V)|<p)>1-2 \textit{ exp}(-2np^2).
\end{equation*}
In this inequality, $\overline{V}$ denotes the empirical estimate of the validation error, $\mathbb{E}[V]$ is its hypothetical true
value, $p$ is the desired maximum deviation between the estimation and the true value, and $n$ represents the
number of samples in the validation set. Based on this inequality, the number of samples in the validation set should be very large. For example, we will require about 20,000 samples in the validation set if we want to be 95\% confident in estimation of validation error that differes less than 1\% from the absolute true value. Note that in this analysis, validation error is computed  as the average of independent binary indicator variables representing  if a given sample in the validation set is classified correctly or not.  This  analysis  may be unrealistic because of the assumption that the validation accuracy is
the average of independent variables. To address this problem, Oliver \textit{et al.} \cite{oliver2018realistic}  measure this phenomenon empirically, and train the SSL methods using 1,000 labels in the  training set from SVHN dataset and then evaluate them on the validation sets with different sizes. Note that these small synthetic  validation sets are generated by different randomly sampled sets without
overlapping from the full SVHN validation set. Following the same setting for evaluation of our SSL algorithm (ROT) in a real world scenario, in Fig. \ref{fig3a} and Fig. \ref{fig4a},  we reported the mean and standard deviation of  validation errors over five times randomly non-overlapping splitting the SVHN and CIFAR validation sets with varying sizes. The results in Fig. \ref{fig3a} and Fig. \ref{fig4a} indicate that as we increase the size of validation set, the  ROT algorithm will be more confident and stable to select its  hyperparameters than the case where we use small-size validation set. For a fair comparison between our method and the other SSL methods in Table. 1 of the paper, we have been consistent with other methods in the size of the training and validation sets as it is designed in standard SVHN and CIFAR-10 datasets.  Specifically, for SVHN, we used  65,932 images for the training set and 7,325 for the validation set, and for CIFAR-10 dataset, we used  45,000 images for the training set and 5,000 images for the validation set. Fig. \ref{fig3b} and Fig. \ref{fig4b} indicate the  error rate of the ROT algorithm on the  SVHN and CIFAR validation sets for different values of $\lambda$ in our transportation plan. Note that during the tuning of $\lambda$, we fixed $\alpha$ in Eq. (5) to one and changed $\lambda$ to different values including 0.1, 0.25, 0.5 , 0.75 and 1. Moreover, during the tuning of $\alpha$, we  fixed $\lambda$ in Eq. (3) to 0.25 and changed $\alpha$ to different values including 0.1, 0.25, 0.5 , 0.75 and 1.  Fig. \ref{fig3c} and Fig. \ref{fig4c} show the  error rate of the ROT algorithm on the SVHN and CIFAR validation sets for different values of $\alpha$ for training the parameters of the  CNN. 
\end{document}